\documentclass[
 showpacs
,showkeys
,twocolumn
,superscriptaddress
]
{revtex4-2}

\usepackage{newfloat,algcompatible}
\usepackage{etoolbox}

\usepackage{graphicx,subfigure,bbm}

\usepackage{amsmath}
\usepackage{booktabs}
\usepackage{algorithm}
\usepackage{algpseudocode}
\usepackage{amsthm}
\usepackage{amssymb}
\usepackage{bm}
\usepackage{flushend}
\usepackage{graphicx}
\usepackage{fancyhdr}
\usepackage{color}
\usepackage{extarrows}
\usepackage{enumerate}
\usepackage{enumitem}
\usepackage{epstopdf}
\usepackage[colorlinks,
            linkcolor=black,
            citecolor=black,
            urlcolor=blue
            ]{hyperref}
\usepackage{tikz}
\usetikzlibrary{shapes,arrows}
\usepackage{soul}
\usepackage{braket}     
\usepackage{physics} 
\usepackage{minted}

\tikzstyle{decision} = [diamond, draw, fill=blue!20,
    text width=4.5em, text badly centered, node distance=3cm, inner sep=0pt]
\tikzstyle{block} = [rectangle, draw, fill=blue!20,
    text width=10em, text centered, rounded corners, minimum height=4em]
\tikzstyle{line} = [draw, -latex']
\tikzstyle{cloud} = [rectangle, draw,fill=red!20, node distance=7cm,
    minimum height=4em]

\begin{document}

\title{Discrete Diffusion Models via Evolving Variational Autoregressive Networks}

\author{Kewen Pan}
\affiliation{Institute of Fundamental and Frontier Sciences, University of Electronic Science and Technology of China, Chengdu 611731, China}

\author{Ying Tang}
\email[Corresponding authors: ]{jamestang23@gmail.com}
\affiliation{Institute of Fundamental and Frontier Sciences, University of Electronic Science and Technology of China, Chengdu 611731, China}
\affiliation{School of Physics, University of Electronic Science and Technology of China, Chengdu 611731, China}
\affiliation{Key Laboratory of Quantum Physics and Photonic Quantum Information, Ministry of Education, University of Electronic Science and Technology of China, Chengdu 611731, China}
\affiliation{Non-classical Information Science Basic Discipline Research Center of Sichuan Province, University of Electronic Science and Technology of China, Chengdu 611731, China}

\begin{abstract}
Conventional score-based diffusion models learn scores without representing normalized densities, whereas tractable normalized models support both sampling and direct likelihood evaluation. A recent tensor-network approach provides such a representation but is largely restricted to low-dimensional lattices. Here we introduce a discrete diffusion model that parameterizes normalized probability distributions using variational autoregressive networks. Explicit Markov jump operators govern the forward noising and reverse denoising dynamics, extending discrete diffusion models with normalized distributions to spin systems on higher-dimensional lattices. We apply this framework to the two- and three-dimensional Ising models across ordered, critical, and disordered regimes, accurately computing thermodynamic quantities including free energy, energy, and magnetization. We further integrate the framework with Monte Carlo sampling, using adaptive diffusion steps to maintain high acceptance rates even at low temperatures while enhancing sample diversity. These results establish a neural-network framework for the discrete diffusion model with normalized probability distributions.
\end{abstract}

\keywords{Discrete diffusion model, Markov chain Monte Carlo, Stochastic dynamics}

\pacs{}

\maketitle

\section{Introduction}
Inspired by non-equilibrium thermodynamics, diffusion models constitute a powerful class of generative models \cite{sohl2015deep}. They follow a two-stage pipeline in which a forward process gradually corrupts samples from the target distribution toward a tractable prior (Uniform, Gaussian, etc.), and a trained reverse process removes this corruption to generate new samples \cite{bahri2020statistical, yang2023diffusion, ho2020denoising}. Nevertheless, many classical diffusion models were originally designed for continuous-valued data, such as images \cite{ho2020denoising, batzolis2021conditional, ho2022cascaded} and audio \cite{schneider2023archisound, lemercier2025diffusion}, and cannot be applied directly to discrete-state systems. 

To adapt to different data modalities, discrete diffusion models corrupt their inputs through explicit state transitions rather than additive Gaussian noise \cite{austin2021structured}, and they have achieved substantial progress in numerous practical domains, such as natural language processing \cite{lou2023discrete, arriola2025block, yu2025discrete, kathuria2026leveraging, wang2026commit, zhou2026steering}, biological sequence generation \cite{gruver2023protein, sarkar2024designing, desurvey, wang2025fine}, and the study of spin systems \cite{sanokowski2025scalable, singh2026independent, zhu2026mdns}. Nevertheless, the most widely used diffusion models are score based: they learn the gradient of the log density and therefore represent only unnormalized densities. This creates three difficulties, which are especially challenging in statistical physics. First, learning the score over the full support of a high-dimensional, sparsely sampled distribution is hard; second, reaching the noise prior requires evolving the forward process for a long time, whereas finite-time denoising in practice introduces reconstruction error \cite{de2021diffusion}; and third, evaluating the likelihood---and hence the free energy---of generated samples from a learned score function is non-trivial.

Recent work on discrete diffusion models based on tensor networks directly confronted these limitations \cite{causer2025discrete}. By encoding the normalized distribution as a matrix product state (MPS) and the diffusion operators as matrix product operators, this approach obtains both a normalized distribution and an exact likelihood—precisely the quantities that score-based models cannot provide. This is an important advance that we build upon. Its principal limitation, however, is representational: MPS is intrinsically adapted to quasi-one-dimensional geometries, and the associated computational cost grows prohibitively with the spatial dimension. Indeed, on moderately sized two-dimensional cylindrical Ising models ($8\times30$ and $9\times30$) the tensor-network magnetization per site already deviates from Monte Carlo benchmarks at the critical temperature, and a direct application to three-dimensional lattices is not feasible. These gaps motivate a normalized discrete diffusion model whose representation is not constrained by the lattice dimension. 

\begin{figure*}
    \centering
    \includegraphics[width=0.75
    \linewidth]{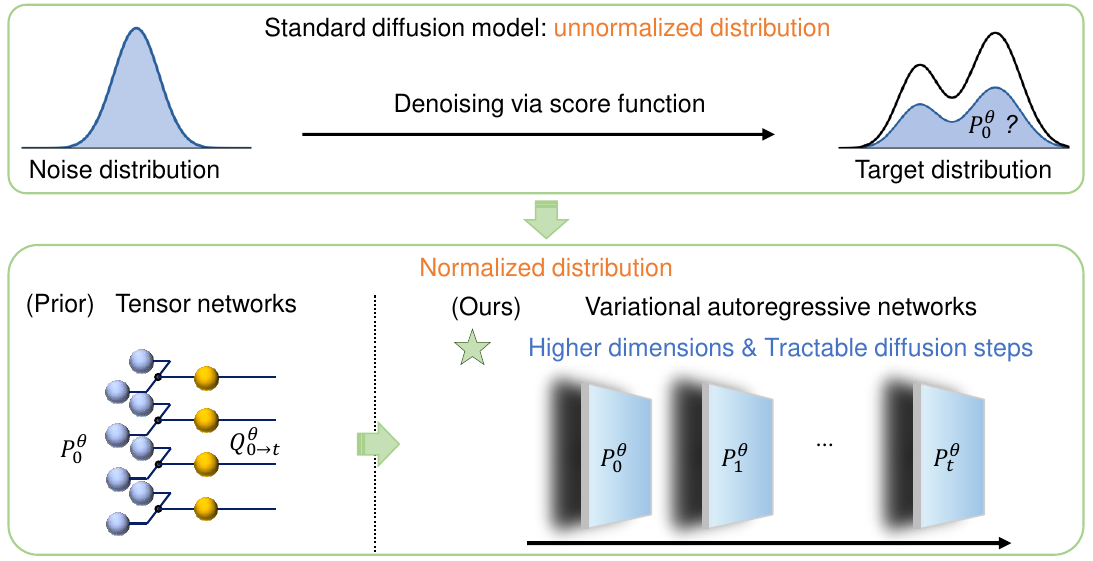}
    \caption{\textbf{Discrete diffusion model by denoising normalized distribution of variational autoregressive networks.} Standard diffusion models adopt a forward noising process and recover samples via a learned score function, without yielding explicit and tractable normalized probability representations for physical simulations. Prior work based on tensor networks addressed this by encoding normalized distributions as matrix product states, and representing the operators acting on them as matrix product operators. Building on that idea, we replace tensor networks with VAN, which learns the normalized distribution at each diffusion step under a fixed transition operator. Compared with tensor networks, our VAN method achieves improved performance in high-dimensional systems and makes diffusion steps tractable.}
    \label{main fig 1.1}
\end{figure*}

In this work, we replace tensor networks with variational autoregressive networks (VAN) \cite{wu2019solving,zhong2026scalable}, an approach that has proven effective for problems in chemical reaction networks \cite{tang2023neural,weng2025tracking}, computational biology \cite{shin2021protein}, and quantum many-body systems \cite{carleo2017solving, sharir2020deep, luo2022autoregressive}. Fig.~\ref{main fig 1.1} explains our research motivation and depicts the framework of our VAN‑based discrete diffusion model. Unlike conventional approaches, which learn score functions and thus represent only unnormalized densities, and unlike the matrix-product-state construction, whose representation is tied to quasi-one-dimensional geometries, our framework retains an explicitly normalized distribution while using an autoregressive ansatz in place of a matrix product state. This yields three concrete gains: (i) the likelihood is available at every diffusion step, so the variational free energy can be evaluated directly; (ii) the ansatz is agnostic to the lattice geometry, so the same network applies to two- and three-dimensional lattices alike, without introducing a bond dimension. We validate the precise modeling capability of our method for spin systems through experiments on 2D and 3D Ising models \cite{chandler1987introduction}, benchmarking thermodynamic observables against tensor network baselines and Wolff cluster Monte Carlo across ordered, critical, and disordered temperature regimes; and (iii) meanwhile, machine-learning-augmented Monte Carlo methods have attracted extensive research attention in recent statistical physics work \cite{causer2025discrete, del2025performance, del2026demonstrating}. Adopting the same sampling paradigm utilized in the tensor network approach, we further integrate our VAN-based discrete diffusion model with Markov chain Monte Carlo (MCMC), which produces non-local updates that break the locality bottleneck of single-spin-flip sampling, sustains high acceptance rates down to low temperatures, and enhances sample diversity.

\begin{figure*}
    \centering
    \includegraphics[width=0.85
    \linewidth]{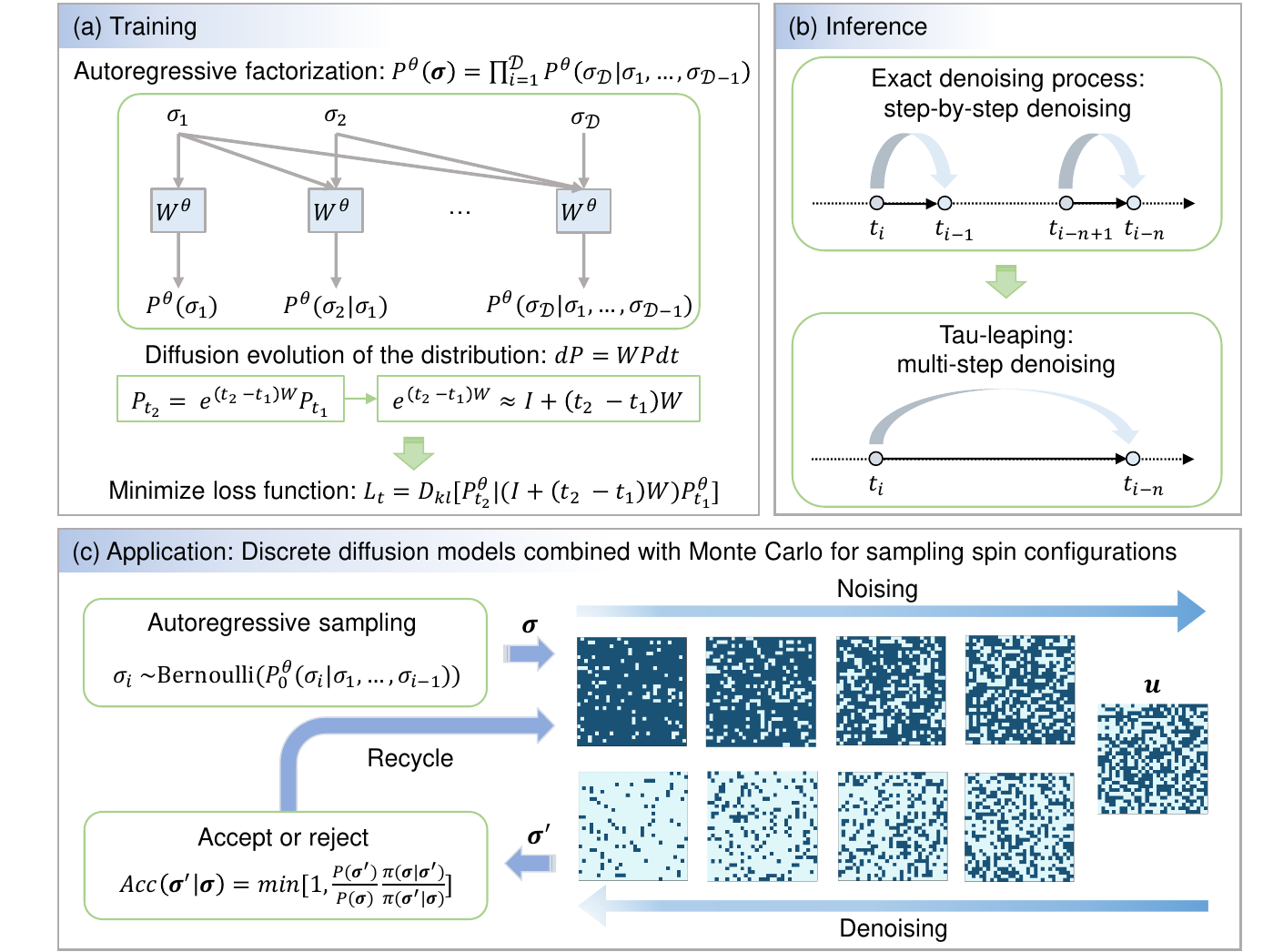}
    \caption{\textbf{Overview of our discrete diffusion model: training procedure, inference strategy, and application.} (a) The autoregressive mechanism and diffusion evolution enable tractable optimization of the loss function. (b) We consider two denoising strategies for inference: an exact denoising scheme for the Euler‑discretized forward noising process, and an accelerated denoising scheme based on the tau-leaping algorithm.  (c) We integrate the discrete diffusion model with the MCMC sampler to enable efficient sampling of spin configurations.}
    \label{main fig 1.2}
\end{figure*}

\section{Background}

\subsection{Variational autoregressive networks}
We consider the problem of sampling the target distribution $P$ defined over a finite discrete state space $\mathcal{S} = \{-1, +1\}^\mathcal{D}$. Specifically, the target distribution adopted in our study follows the Boltzmann distribution
\begin{equation}
P(\boldsymbol{\sigma}) = \frac{1}{Z} e^{ - \beta E(\boldsymbol{\sigma})},
\label{eq:Boltzmann Distribution}
\end{equation}
where \(\beta = 1/T\) denotes the inverse temperature, \(E(\boldsymbol{\sigma})\) is a known energy function, and \(\boldsymbol{\sigma} = (\sigma_1, \sigma_2, \dots,\sigma_{\mathcal{D}})\) denotes the spin configuration, with each discrete spin variable satisfying \(\sigma_i \in \{-1, +1\}\). The term \(Z\) denotes the partition function, a critical normalization constant defined as the sum of exponential energy terms over all configurations of the system
\begin{equation}
Z = \sum_{\boldsymbol{\sigma}} e^{-\beta E(\boldsymbol{\sigma})}.
\label{partition function}
\end{equation}
The spin system has a state space containing \(2^\mathcal{D}\) distinct configurations, so the number of configurations grows exponentially with the system size \(\mathcal{D}\). This exponential scaling of the configuration space means that the exact numerical computation of $Z$ becomes computationally intractable for systems with large \(\mathcal{D}\).

To address this limitation, the VAN achieves tractable modeling via a variational autoregressive factorization scheme in Fig.~\ref{main fig 1.2}(a), which decomposes the intractable joint probability distribution into a product of conditional probabilities in a prescribed sequential order
\begin{equation}
P^\theta(\boldsymbol{\sigma}) = \prod_{i=1}^\mathcal{D} P^\theta(\sigma_i \mid \sigma_1, \dots, \sigma_{i-1}),
\label{eq:product of conditional probabilities}
\end{equation}
where $\theta$ denotes the learnable parameters. Owing to its inherent autoregressive architecture, the VAN possesses two notable features for the Boltzmann distribution approximation. First, it is capable of capturing long-range spatial correlations among spin variables. Second, the sequential factorization structure supports precise likelihood evaluation of generated configurations \(P^\theta(\boldsymbol{\sigma})\) and autoregressive sampling, allowing independent configuration generation in a predefined order.

The training objective of VAN is to optimize the parameterized variational ansatz $P^\theta$ to precisely fit the target Boltzmann distribution $P$. To quantitatively measure distribution mismatch and guide the optimization process, we utilize the Kullback-Leibler (KL) divergence, a standard measure for evaluating the difference between two probability distributions. Specifically, we employ reverse KL divergence, which only requires explicit knowledge of the energy function and allows model training using samples drawn directly from the model itself:
\begin{align}
\label{Kullback-Leibler (KL) divergence}
D_{\text{KL}}(P^\theta||P) 
&= \sum_{\boldsymbol{\sigma}} P^\theta(\boldsymbol{\sigma}) \log \left( \frac{P^\theta(\boldsymbol{\sigma})}{P(\boldsymbol{\sigma})} \right)\notag \\
&= \beta (F(\theta) - F),
\end{align}
where 
\begin{align}
\label{Free energy}
F = -\frac{1}{\beta}\log Z
\end{align}
represents the free energy of the system. Owing to the non-negativity of the KL divergence, the variational free energy
\begin{align}
\label{variational free energy}
F(\theta)=\frac{1}{\beta} \sum_{\boldsymbol{\sigma}} P^\theta(\boldsymbol{\sigma}) \left( \log P^\theta(\boldsymbol{\sigma}) + \beta E(\boldsymbol{\sigma}) \right)
\end{align}
serves as an upper bound of the exact free energy $F$. Minimizing \(F(\theta)\) narrows the gap between this variational free energy \(F(\theta)\) and the exact free energy $F$, while driving the parameterized distribution \(P^\theta(\boldsymbol{\sigma})\) to gradually converge to the exact distribution of the system.

Compared to tensor networks \cite{levin2007tensor, xie2009second}, the key advantage of VAN is the ability to approximate complex target distributions with its superior generality. Unlike MPS, which are inherently limited to one-dimensional systems, the neural network-based ansatz of VAN can be applied to systems with arbitrary topological structures. For example, convolutional neural networks (CNN) \cite{ketkar2021convolutional} are highly suitable for modeling two-dimensional systems, as their convolutional architecture efficiently captures local spatial structure in 2D data, similar to how it processes image information.

VAN has proven to be a powerful tool for studying equilibrium statistical mechanics \cite{mehta2019high, mcnaughton2020boosting, biazzo2024sparse}. Beyond equilibrium problems, it is also applicable to exploring the time evolution dynamics of nonequilibrium systems in statistical mechanics \cite{tang2024learning}. By accurately approximating the normalized probability distribution at each time step, the VAN enables efficient calculation of the dynamical partition function — a core quantity for characterizing dynamical phase transitions and extracting essential dynamical observables. This capability broadens the research scope of statistical mechanics, allowing researchers to reveal finite-time dynamical phase diagrams, critical scaling laws, and emergent spatial structures that are challenging for conventional analytical and numerical approaches. Discrete diffusion models are designed to describe unidirectional stochastic Markov processes and thus belong to the class of nonequilibrium dynamical systems. Leveraging the strong capability of VAN in capturing nonequilibrium evolution, we integrate the above VAN-based dynamic learning framework into discrete diffusion models.

\subsection{Discrete noising process}
We formulate the noising dynamics as a continuous time Markov chain (CTMC), which is a stochastic process on a discrete state space $\mathcal{S}$ with the Markov property. The time evolution of the probability distribution $P_t$ satisfies the master equation
\begin{align}
\label{master equation}
\frac{dP_t}{dt} = W P_t,
\end{align}
where the standard form of the Markov generator $W$ is given by
\begin{align}
\label{master equation generator}
W_{\boldsymbol\sigma'\boldsymbol\sigma} =
\begin{cases}
w(\boldsymbol\sigma' \leftarrow \boldsymbol\sigma), & \text{if } \boldsymbol\sigma' \neq \boldsymbol\sigma \\
-R(\boldsymbol\sigma), & \text{if } \boldsymbol\sigma' = \boldsymbol\sigma.
\end{cases}
\end{align}

We consider the bistochastic noising dynamics built from non-interacting single-spin flips. Transitions are allowed only between configurations separated by a Hamming distance \(d=1\), i.e., configurations differing by a single spin flip. The transition rates are defined as
\begin{align}
\label{transition rates}
w(\boldsymbol\sigma' \leftarrow \boldsymbol\sigma) = 
\begin{cases}
1, & d(\boldsymbol\sigma', \boldsymbol\sigma) = 1 \\
0, & \text{otherwise}.
\end{cases}
\end{align}
The corresponding escape rates are $R(\boldsymbol\sigma) = \sum_{\boldsymbol\sigma' \neq \boldsymbol\sigma} w(\boldsymbol\sigma' \leftarrow \boldsymbol\sigma) = \mathcal{D}.$ This construction endows $W$ with two key properties: probability normalization is preserved under time evolution, and the probability distribution asymptotically converges to the uniform distribution, which serves as the stationary distribution satisfying \(W P_{\text{uniform}} = 0\).

Based on the master equation Eq.~\eqref{master equation}, the evolution of the probability distribution from \(t = t_1\) to \(t = t_2\) can be equivalently expressed as
\begin{align}
    {P_{t_2}^\theta} = e^{(t_2 - t_1) W} {P_{t_1}^\theta} 
    = {Q}_{t_2 \leftarrow t_1} {P_{t_1}^\theta}.
    \label{eq:evolution}
\end{align}
For a sufficiently small time interval \(\Delta t = t_2 - t_1\), the matrix can be approximated via the first‑order Euler expansion: \(e^{\Delta t W} \approx \mathbb{I} + \Delta t W\), where \(\mathbb{I}\) denotes the identity matrix and $\mathcal{D}\Delta t \leq 1$. A typical choice satisfying this bound is \(\Delta t = 1/(2\mathcal{D})\). Accordingly, the transition probability under this Euler approximation takes the form
\begin{align}
\label{transition probability}
p_{t_2|t_1}(y|x) = \delta(y,x)+\Delta tW(y,x)+o(\Delta t),
\end{align}
where $o(\Delta t)$ represents higher-order infinitesimal terms vanishing faster than $\Delta t$, and $\delta(y,x)$ is the Kronecker delta that takes value 1 when $x=y$ and 0 otherwise. We adopt this first‑order Euler approximation to construct the forward noising process of our discrete diffusion model, whose dynamics are governed by the single‑spin‑flip generator $W$.

\subsection{Discrete denoising process}
Efficient simulation of discrete stochastic processes is a fundamental challenge in statistical physics and computational modeling. For our discrete diffusion model, which enables tractable probability flow between adjacent time steps, the Euler-discretized reverse process can be analytically derived based on Bayes’ theorem \cite{sun2022score}
\begin{align}
\pi_{t_1 \mid t_2}(x_1 \mid x_2) = \pi_{t_2 \mid t_1}(x_2 \mid x_1) \frac{P_{t_1}(x_1)}{P_{t_2}(x_2)},\quad t_1 < t_2.
\label{Bayes’ theorem}
\end{align}
The Alg.~\ref{alg:exact_denoising} details the full implementation pipeline for the Euler-discretized denoising update rules of our discrete diffusion model.

\begin{algorithm}[H]
\caption{Stepwise Euler-discretized denoising}
\label{alg:exact_denoising}
\begin{algorithmic}[1]
\small
\Require A transition rate $w$, time partition $0 = t_0 < t_1 < \cdots < t_K$.
\State Draw $u_K \sim P_{t_K}^{\theta}$.
\For{$k = K$ \textbf{downto} $1$}
    \State Set $u_{k-1} = $
    \State \quad $\begin{cases}
    u, & \text{w.p. } \Delta t_k w\frac{P_{t_{k-1}}^{\theta}(u)}{P_{t_k}{^{\theta}(u_k)}} \\[3pt]
    u_k, & \text{w.p. } 1 - \sum\limits_{u\neq u_k} \Delta t_k w\frac{P_{t_{k-1}}^{\theta}(u)}{P_{t_k}{^{\theta}(u_k)}}
    \end{cases}$
    \State \quad where $\Delta t_k = t_k-t_{k-1}$ and $u\neq u_k$.
\EndFor
\State \Return $u_0 \sim P_{t_0}^{\theta}$.
\end{algorithmic}
\end{algorithm}

The stepwise Euler-discretized denoising method takes the form of a two-stage algorithm outlined below:
\begin{enumerate}[label=(\roman*), leftmargin=*, itemsep=0.1em]
    \item Firstly, it initializes the noisy states $u_K$. The states are generated either by applying sequential flips to the $t=0$ samples following our noising rules or by direct sampling from the distribution $P_{t_K}^{\theta}$.
    \item Then it iteratively reconstructs the latent states $u_{k-1}$ from $u_k$. The probability of updating $u_k$ is given by $\Delta t_k w P_{t_{k-1}}^{\theta}(u)/P_{t_k}^{\theta}(u_k)$. Instead of directly evaluating \(P_{t_k}^{\theta}(u_k)\), we can compute it from \(P_{t_{k-1}}^{\theta}\) using the noising rule to ensure that the probability of retaining the state \(u_k\) remains positive.
\end{enumerate}

Although this stepwise denoising approach yields accurate results, it suffers from substantial computational overhead and low sampling efficiency, particularly for long denoising trajectories. To balance numerical accuracy and computational cost, we incorporate the tau‑leaping into our discrete diffusion model.

Tau-leaping is an approximate stochastic simulation method designed to accelerate sampling of discrete-state Markov processes, offering noticeable efficiency improvements for systems with frequent state transitions.  Originally proposed and widely employed in chemical physics, this technique accelerates simulations of large reaction networks with abundant reaction events \cite{gillespie2001approximate, cao2005avoiding, cao2006efficient}.  More recently, tau-leaping has been extended to discrete diffusion models based on CTMC for an efficient approximation of the generative reverse process \cite{campbell2022continuous}. Unlike the classic Gillespie algorithm \cite{gillespie1976general, gillespie1977exact}, which simulates only one transition event per iteration, tau-leaping bundles numerous transition events in a single step to reduce computational burden. This property makes tau-leaping particularly suitable for high-dimensional systems with rapid transition rates.

\begin{algorithm}[H]
\caption{Tau-leaping}
\label{Alg:tau-leaping}
\begin{algorithmic}[1]
\Require A transition rate $w$, time partition $0 = t_0 < t_1 < \cdots < t_K$, optimal leap duration $\tau$.
\State Draw $u_K \sim P_{t_K}^{\theta}$, where $K = \lceil \dfrac{t_K}{\tau} \rceil$.
\State Encoding map: $\phi(u_K)={(u_K+1)}/{2}$
\For{$k = K$ \textbf{downto} $1$}
    \For{$d = 1$ \textbf{to} $\mathcal{D}$} 
        \For{$z^d \in \{0,1\} \setminus \{u_{k}^d\}$}
            \State Draw $M_{d,z^d} \sim \text{Poisson}(\tau w P_{t_k}^{\theta}(u_k^{d,z^d})/P_{t_k}^{\theta}(u_k))$.
        \EndFor
    \EndFor
    \State Set $x = u_{k} + \sum_{d=1}^{\mathcal{D}} \sum_{z^d \in \{0,1\} \setminus \{u_{k}^d\}} (z^d - u_{k}^d) M_{d,z^d}$.
    \State Set $u_{k-1} = \mathcal{M}_{u_{k}}(x)$, where $\mathcal{M}_{u_{k}}(x) = \{x^d \delta_\mathcal{S}(x^d) + z^d(1 - \delta_\mathcal{S}(x^d))\}_{d \in [\mathcal{D}]}$ is a mapping from $\mathbb{Z}^\mathcal{D}$ to $\mathcal{S}^\mathcal{D}$.
\EndFor
\State \Return Decoding map: $\phi^{-1}(u_0)={2\,u_0-1}$
\end{algorithmic}
\end{algorithm}

Compared with the Alg.~\ref{alg:exact_denoising}, the tau-leaping method for our discrete diffusion model also adopts a two-stage procedure and retains an identical first stage. In its second stage, stepwise Euler-discretized denoising is substituted with an approximate leap over a predetermined optimal leap duration \(\tau\) (Supplementary Appendix~\ref{Optimal leap}). Instead of simulating every transition separately, the tau-leaping method assumes that the reverse transition rates and the state $u_k$ remain constant throughout the interval \(\tau\). We sample Poisson counts $M_{d,z^d}$ for each coordinate $d$, aggregate these counts and apply them to the state \(u_k\) at the end of the interval \(\tau\) to form the state $x$. $x$ is then projected onto the valid discrete state space \(\mathcal{S}\) via the mapping \(\mathcal{M}_{u_k}(\cdot)\) to produce the latent state \(u_{k-1}\). Although the tau-leaping method reduces computational cost, we revert to the stepwise Euler-discretized denoising method when strict numerical accuracy is required. 

\section{Numerical results}
\begin{figure*}[t]
    \centering
    \includegraphics[width=0.85
    \linewidth]{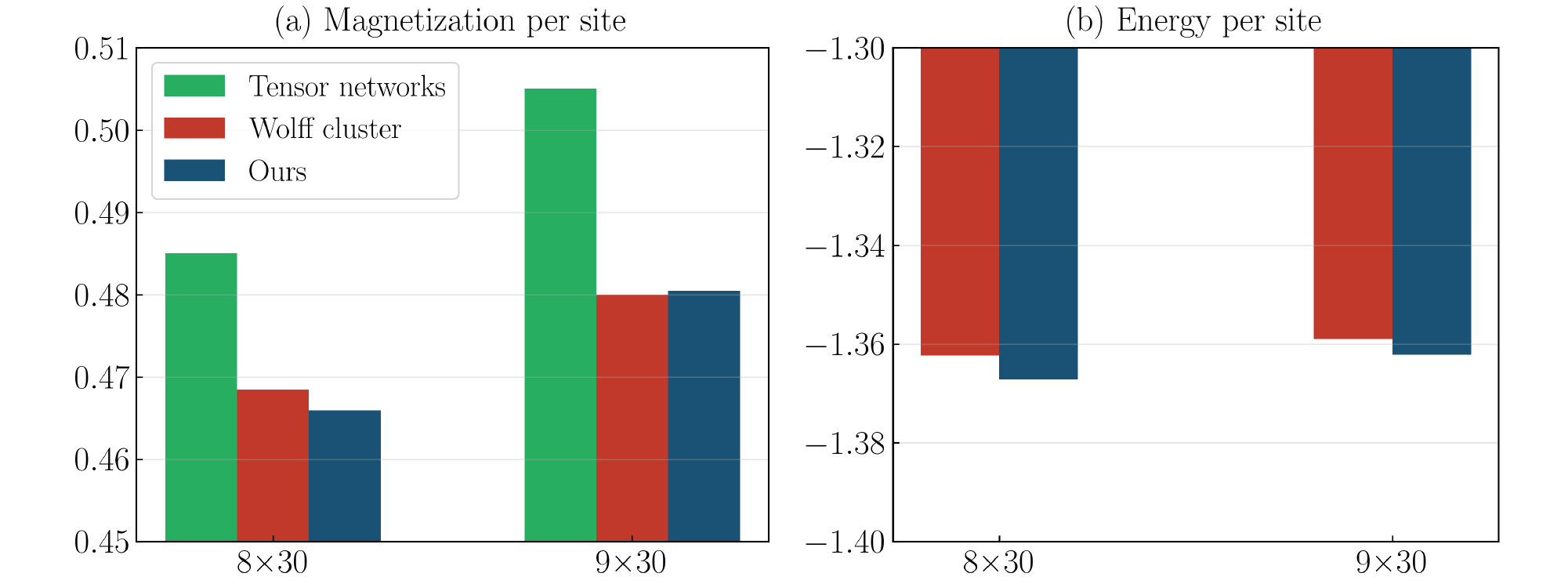}
    \caption{\textbf{Tensor networks and VAN-based discrete diffusion models on 2D Ising model.} (a) Magnetization per site obtained via the tensor network approach \cite{causer2025discrete}, the Wolff cluster algorithm, and our method with averages taken over \(10^5\) independent samples, on \(8 \times 30\) and \(9 \times 30\) lattices. (b) Comparison of energy per site under the same simulation parameters. Note that energy per site data from the tensor network approach are unavailable in the original reference.}
    \label{compare 2D}
\end{figure*}
The Ising model is defined on a discrete spin lattice with nearest-neighbor pairwise interactions, whose Hamiltonian is given by 
\begin{align}
\label{energy function}
E(\boldsymbol{\sigma}) = -J\sum_{\langle i,j \rangle} \sigma_i \sigma_j, 
\end{align}
where $\langle i,j \rangle$ denotes nearest-neighbor pairs of spins, and the ferromagnetic coupling strength is set to $J=1$. In the thermodynamic limit, the two-dimensional Ising model undergoes a continuous phase transition at the critical inverse temperature $\beta_c = \ln(1+\sqrt{2}) / 2$. The system remains disordered for $\beta < \beta_c$ at high temperatures and develops spontaneous magnetic order for $\beta > \beta_c$ at low temperatures.

We first consider the two-dimensional Ising model defined on a \(\mathcal{D} = L_1 \times L_2\) lattice (\(L_2 \geq L_1\)) with cylindrical boundary conditions, where periodic boundary conditions (PBC) are applied along the first dimension and open boundary conditions (OBC) along the second dimension. We adopt lattice sizes of $8\times30$ and $9\times30$ at the critical inverse temperature $\beta_c$, which produce numerical biases in the tensor network approach compared to standard Monte Carlo benchmarks \cite{causer2025discrete}. To quantify the accuracy of the learned distributions, we define the absolute magnetization per site as 
\begin{align}
\label{absolute magnetization per site}
m = \left| \frac{1}{\mathcal{D}}\sum_{i=1}^\mathcal{D} \sigma_i \right|,
\end{align}
and compare both the magnetization and the energy per site against the reference data. The comparison results in Fig.~\ref{compare 2D} demonstrate that our VAN method achieves higher accuracy in predicting magnetization per site than tensor networks, while producing reliable estimates for the energy per site. We adopt results from the Wolff cluster algorithm (Supplementary Alg.~\ref{Wolff Cluster}) as reference data in our work.

\begin{figure*}[t]
    \centering
    \includegraphics[width=1
    \linewidth]{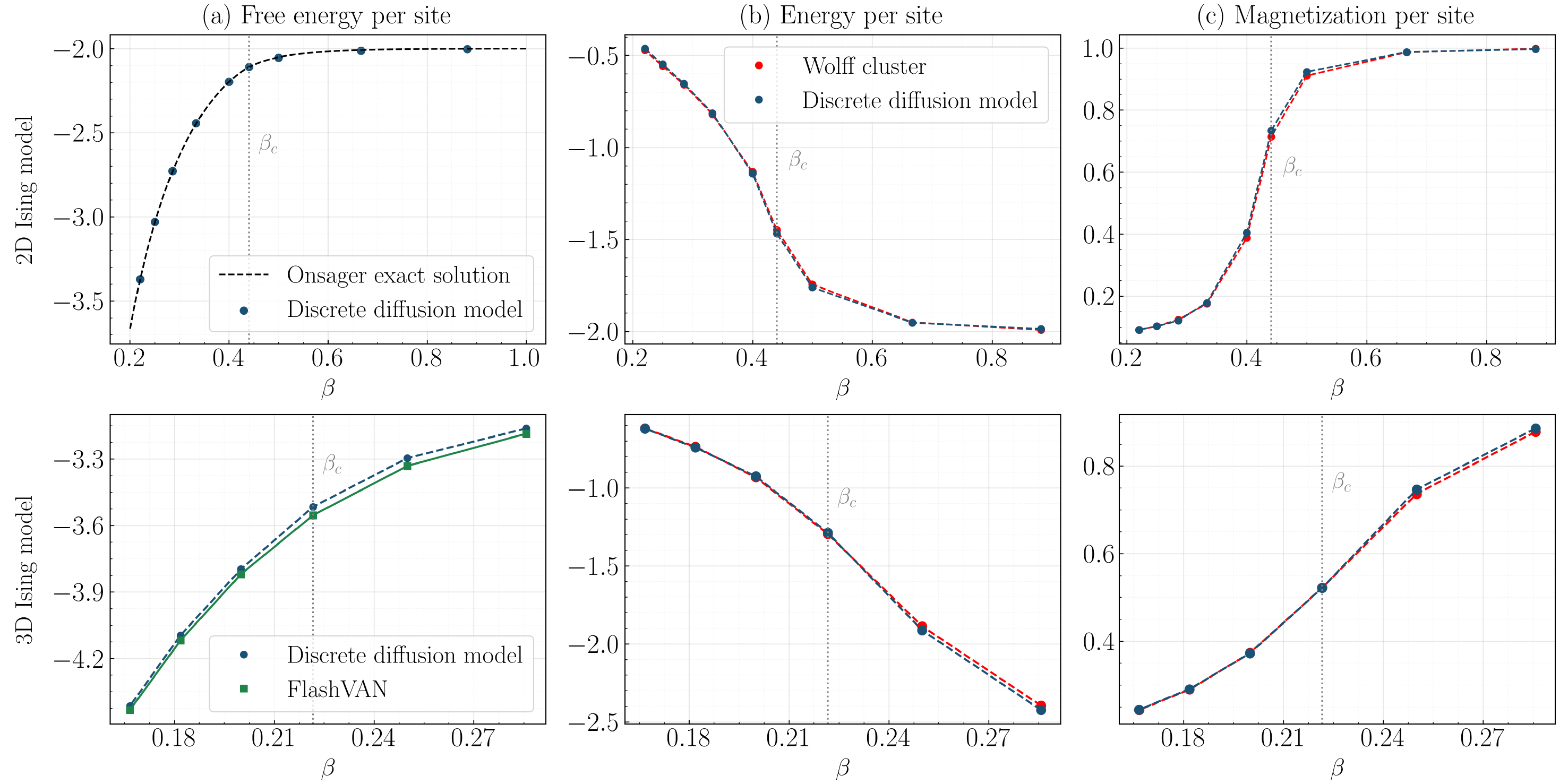}
    \caption{\textbf{Learned discrete diffusion models for 2D and 3D Ising models at distinct temperatures.} (a) Free energy per site, (b) energy per site and (c) magnetization per site as functions of inverse temperature $\beta$. The upper row corresponds to the $\mathcal{D}=16\times16$ two-dimensional Ising model with temperatures $T = 1/\beta = 1/\ln(1+\sqrt{2}), 1.5000, 2.0000, 2/\ln(1+\sqrt{2}), 2.5000, 3.0000, 3.5000, 4.0000, 4/\ln(1+\sqrt{2})$. The lower row presents statistical results for the $\mathcal{D}=4\times4\times4$ three-dimensional Ising model at temperatures $T = 1/\beta = 3.5000, 4.0000, 4.5115, 5.0000, 5.5000, 6.0000$. All systems are simulated with periodic boundary conditions and all results are averaged over $10^5$ independent samples.}
    \label{2D 3D ising}
\end{figure*}

We further perform systematic simulations on both 2D and 3D Ising models with full PBC: for 2D lattices, all horizontal and vertical boundaries are periodically connected; for 3D lattices, periodicity is imposed along all three spatial dimensions, which guarantees uniform nearest-neighbor spin interactions across the entire system. We compute the free energy per site, energy per site, and magnetization per site for lattices of size $16\times16$ (2D) and $4\times4\times4$ (3D) over a wide temperature range spanning the low-temperature ordered regime, the high-temperature disordered regime, and the phase transition point. Three-dimensional lattices are substantially more demanding for tensor networks, whose computational cost and memory usage grow exponentially with the spatial dimension and severely hinder the accurate characterization of long-range spin correlations. This is precisely the regime in which a neural autoregressive representation remains tractable, and the resulting estimates are summarized in Fig.~\ref{2D 3D ising}. Since the 3D Ising model has no exact solution for free energy, we computed the free energy using FlashVAN \cite{zhong2026scalable}—currently the state-of-the-art method—and used the resulting values as a benchmark for comparison.

\section{Integrating discrete diffusion into MCMC proposal updates}
From a theoretical perspective, perfect alignment between the learned VAN $P^\theta$ and the target equilibrium distribution would render autoregressive sampling fully unbiased, eliminating the need for supplementary MCMC correction. In practice, however, practical neural network approximations inevitably suffer from a distribution mismatch against the true distribution, due to limited training data, inherent architectural constraints of neural networks, and fundamental optimization barriers. A typical symptom of this discrepancy is mode collapse, inducing biased results and poor sample diversity. Furthermore, the learned distribution may correspond to a high temperature distribution, whereas our objective is to sample the target distribution at lower temperatures. Instead of merely retraining the neural network to eliminate such distribution gaps, we introduce a proposal module based on a discrete diffusion model (DDM) embedded in the MCMC workflow, which enables efficient sampling of the desired target distribution \cite{hunt2024accelerating, chen2026markov}. 

To ensure reliable MCMC sampling, we need to train a discrete diffusion model to recover corrupted samples and steer them toward reasonable data distributions. Unlike standard discrete diffusion models, our method adopts a fundamentally different stepwise training scheme for modeling sequential distributions. To adapt to this specialized training paradigm, we first formulate step-by-step loss functions to optimize the VAN parameters
\begin{align}
\label{loss function}
\mathrm{loss}(P_t^{\theta}) = 
\begin{cases}
  \mathbb{E}_{\boldsymbol{\sigma} \sim P_t^{\theta}} \left[ \log P_t^{\theta}(\boldsymbol{\sigma}) + \beta E(\boldsymbol{\sigma}) \right], & t = 0 \\
  \mathbb{E}_{\boldsymbol{\sigma} \sim P_t^{\theta}} \left[ \log P_t^{\theta}(\boldsymbol{\sigma}) - \log P_t(\boldsymbol{\sigma}) \right], & t > 0.
\end{cases}
\end{align}
Concretely, the training pipeline of our discrete diffusion model is implemented in two stages: 
\begin{enumerate}[label=(\roman*), leftmargin=*, itemsep=0.1em]
    \item We first train a VAN to fit the target distribution, which serves as the initial distribution of the diffusion process at step \(t=0\).
    \item Given the trained distribution \(P_{t-1}^{\theta}\), we employ the transition rate $w$ and the time step \(\Delta t\) to compute the supervised training target \(P_t(\boldsymbol{\sigma})\), which is required to train the VAN of the subsequent diffusion step. Iterative execution of this procedure constructs the complete forward noising process in an explicit and deterministic manner. For all \(t>0\), we initialize the VAN with the trained parameters from the previous step and fine-tune it with a small number of optimization iterations to obtain the distribution \(P_{t}^{\theta}\).
\end{enumerate}

Equipped with the trained discrete diffusion model, we construct our adaptive MCMC sampler. Fig.~\ref{main fig 1.2}(c) illustrates its complete framework and pipeline, and each MCMC iteration proceeds as follows.

(1) All MCMC chains are initialized with spin lattice configurations $\boldsymbol{\sigma}$ sampled autoregressively from the VAN-learned distribution $P_0^{\theta}$. Under this autoregressive sampling scheme, the spin variable \(\sigma_i\) at each lattice site $i$ is sampled from its conditional distribution given all previously sampled spins
\begin{align}
\label{VAN sampling}
\sigma_i \sim \mathrm{Bernoulli}\left(P_0^\theta\left(\sigma_i \mid \sigma_1, \dots, \sigma_{i-1}\right)\right).
\end{align}

(2) Then the initial spin configurations $\boldsymbol{\sigma}$ are progressively corrupted through a sequence of forward diffusion steps $t$ to produce noisy states $\boldsymbol{u}$, which disrupt the original structural features of the spin lattice. Starting from these corrupted configurations $\boldsymbol{u}$, we invert the noising process via either the Euler-discretized denoising or the tau‑leaping method to reconstruct candidate spin states $\boldsymbol{\sigma}'$. The corresponding transition probability of this noising-and-denoising proposal process is given by 
\begin{align}
\pi({\boldsymbol{\sigma}}^{\prime} \mid {\boldsymbol{\sigma}}) &= \sum_{\boldsymbol{u}} P_{ t \mid 0}({\boldsymbol{u} \mid \boldsymbol{\sigma}})P_{0 \mid t}({\boldsymbol{\sigma}}^{\prime} \mid \boldsymbol{u})\notag\\
&= \sum_{\boldsymbol{u}} P_{ t \mid 0}({\boldsymbol{u} \mid \boldsymbol{\sigma}})P_{ t \mid 0}({\boldsymbol{u} \mid {\boldsymbol{\sigma}}^{\prime}})\frac{P_0^{\theta}({\boldsymbol{\sigma}}^{\prime})}{P_t^{\theta}(\boldsymbol{u})}.
\label{eq:noising conditional probability}
\end{align}
 
\begin{figure}[t]
    \centering
    \includegraphics[width=1
    \linewidth]{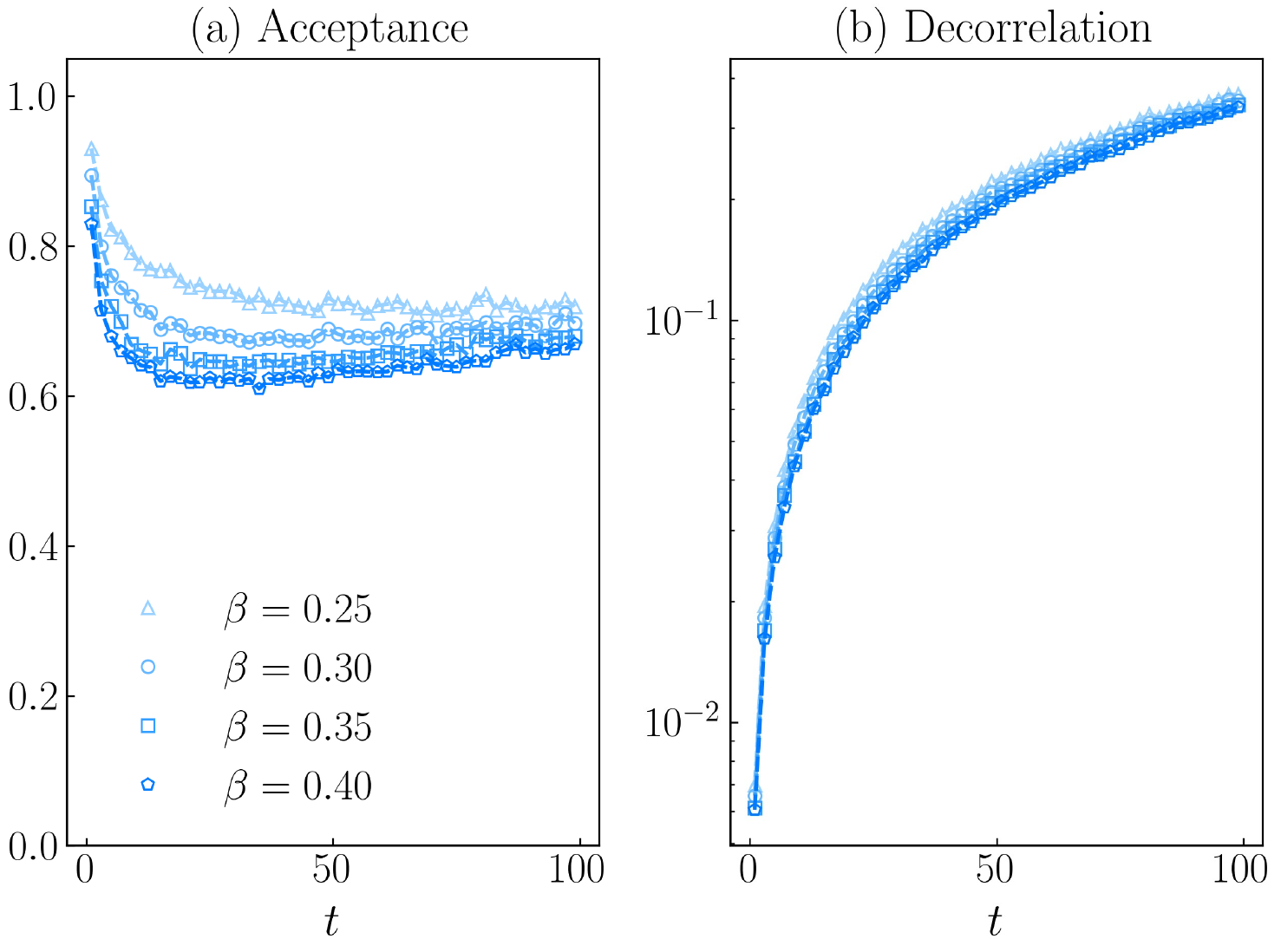}
    \caption{\textbf{Sampling characteristics for different distributions and diffusion steps.} (a) Acceptance rate as a function of diffusion steps for $N = 128$ independent configurations of the $16\times16$ two-dimensional Ising model in the first MCMC iteration, each initialized at inverse temperature $\beta = \beta_c / 2$ and quenched to target inverse temperatures $\beta = 0.25, 0.30, 0.35$ and $0.40$ (repeated over 40 independent runs). (b) Decorrelation coefficient versus diffusion steps for the same set of configurations at each target temperature. The results reveal two consistent trends across all tested temperatures: a larger temperature gap reduces both the acceptance and configuration decorrelation, while increasing the diffusion steps lowers the acceptance via a sharp drop at early steps but improves configuration decorrelation.}
    \label{monte carlo quench-1}
\end{figure}

(3) To draw samples that conform to the target equilibrium distribution, candidate spin configurations \(\boldsymbol{\sigma}'\) are further accepted or rejected according to the standard Metropolis–Hastings acceptance probability, defined as
\begin{align}
\text{Acc}\left[{\boldsymbol{\sigma}}\rightarrow{\boldsymbol{\sigma}}^{\prime}\right]
&=\min\left[1,\frac{P_{\text{true}}({\boldsymbol{\sigma}}^{\prime}) \pi({\boldsymbol{\sigma}} \mid {\boldsymbol{\sigma}}^{\prime})}{P_{\text{true}}({\boldsymbol{\sigma}})\pi({\boldsymbol{\sigma}}^{\prime} \mid {\boldsymbol{\sigma}})}\right]\notag \\
&=\min\left[1,\frac{e^{ - \beta E({\boldsymbol{\sigma}}^{\prime})} P_0^{\theta}({\boldsymbol{\sigma}})}{e^{ - \beta E({\boldsymbol{\sigma}})} P_0^{\theta}({\boldsymbol{\sigma}}^{\prime})}\right],
\label{eq:acceptance}
\end{align}
where the simplification adopts the conditional probability formula in Eq.~\eqref{eq:noising conditional probability}. When employing tau‑leaping for denoising, a sufficient numerical accuracy must be maintained to ensure that Eq.~\eqref{eq:acceptance} is satisfied. In Fig.~\ref{monte carlo quench-1}, we investigate how acceptance and decorrelation vary with diffusion steps and target‑distribution temperature. At fixed diffusion steps, a larger temperature gap lowers acceptance and leads to weaker decorrelation. Interestingly, more diffusion step updates reduce acceptance, but the additional spin flips applied to candidate configurations enhance decorrelation upon acceptance. We define the decorrelation coefficient as \(1 - C\), where \(C\) is the normalized correlation coefficient
\begin{align}
C = \frac{\displaystyle \sum_{d=1}^\mathcal{D} \left( \frac{1}{N-1} \sum_{i=1}^N \left( \sigma_{i,d}^{(k)} - \bar{\sigma}_d^{(k)} \right) \left( \sigma_{i,d}^{(k+1)} - \bar{\sigma}_d^{(k+1)} \right) \right)}{\displaystyle \sqrt{\left( \sum_{d=1}^\mathcal{D} \text{Var}\!\left( \boldsymbol{\sigma}^{(k)}_d \right) \right) \left( \sum_{d=1}^\mathcal{D} \text{Var}\!\left( \boldsymbol{\sigma}^{(k+1)}_d \right) \right)}}
\label{eq:correlation}
\end{align}
to evaluate statistical independence between successive sampled spin configurations $\left(\boldsymbol{\sigma}^{(k)}, \boldsymbol{\sigma}^{(k+1)}\right)$. Here, \(\boldsymbol{\sigma}^{(k+1)}\) denotes the configuration obtained after the Metropolis–Hastings acceptance step. Our adaptive strategy can effectively balance acceptance and decorrelation to enhance the quality of samples: we increase the diffusion time $t$ if the acceptance probability exceeds the preset target threshold and decrease $t$ otherwise. The spin configurations obtained from the previous MCMC iteration are then adopted as initial states for the next iteration, and this sampling pipeline is run iteratively until all Markov chains achieve convergence.

Our MCMC sampler exploits the two-stage noising–denoising dynamics of discrete diffusion models to construct high quality proposal samples for spin lattice systems. In our framework, the forward diffusion process introduces a sequence of unconstrained random spin flips to diversify the lattice structures, which effectively shifts samples to a “higher-temperature” regime with a much more accessible state space for broad exploration \cite{chen2026markov}. The subsequent reverse denoising process imposes structured constraints on the spin flips to guide the empirical distribution of the generated samples to align with the original distribution \(P_0^{\theta}\), reverting configurations to the “lower-temperature” regime. By restricting spin flips, this denoising procedure mitigates large deviations of candidate samples from the target distribution that arise from random spin flips, which moderately increases the acceptance rate in the subsequent Metropolis–Hastings acceptance step. By evolving spin configurations through complete forward corruption and reverse denoising reconstruction, our method generates non-local lattice perturbations, fundamentally breaking the spatial locality bottleneck of the conventional Local Metropolis Monte Carlo (LMMC; Supplementary Alg.~\ref{LMMC}). Traditional LMMC relies on limited local updates, which frequently trap Markov chains in free-energy local minima during long simulations and fail to fully explore the entire configuration space.

Our MCMC method built on discrete diffusion models corresponds to the connected update proposed in prior work \cite{causer2025discrete}, where the current spin configuration $\boldsymbol{\sigma}$ and the proposed candidate configuration $\boldsymbol{\sigma}^\prime$ are statistically correlated. Meanwhile, the prior work also presents a disconnected update scheme, which generates a fully independent candidate configuration $\boldsymbol{\sigma}^\prime$ with no correlation to the original state $\boldsymbol{\sigma}$. In our work, we adopt the neural MCMC (NMCMC) to implement such disconnected updates (Supplementary Alg.~\ref{NMCMC}). The generated candidate configuration is accepted or rejected following the acceptance probability
\begin{align}
\text{Acc}\left[{\boldsymbol{\sigma}}\rightarrow{\boldsymbol{\sigma}}^{\prime}\right]
&=\min\left[1,\frac{P_{\text{true}}({\boldsymbol{\sigma}}^{\prime}) \pi({\boldsymbol{\sigma}} \mid {\boldsymbol{\sigma}}^{\prime})}{P_{\text{true}}({\boldsymbol{\sigma}})\pi({\boldsymbol{\sigma}}^{\prime} \mid {\boldsymbol{\sigma}})}\right]\notag \\
&= \min\left[1,\, \frac{e^{-\beta E(\boldsymbol{\sigma}')}P_0^\theta(\boldsymbol{\sigma})}{e^{-\beta E(\boldsymbol{\sigma})}P_0^\theta(\boldsymbol{\sigma}')}\right],
\label{NMCMC-acc}
\end{align}
where the candidate state \(\boldsymbol{\sigma}'\) is drawn from the marginal proposal distribution $P_0^{\theta}$ parameterized by the VAN. Since \(\boldsymbol{\sigma}'\) is statistically independent of the current configuration \(\boldsymbol{\sigma}\), the transition kernel satisfies \(\pi(\boldsymbol{\sigma}' \mid \boldsymbol{\sigma}) = P_0^\theta(\boldsymbol{\sigma}')\). 

\begin{figure*}
    \centering
    \includegraphics[width=1
    \linewidth]{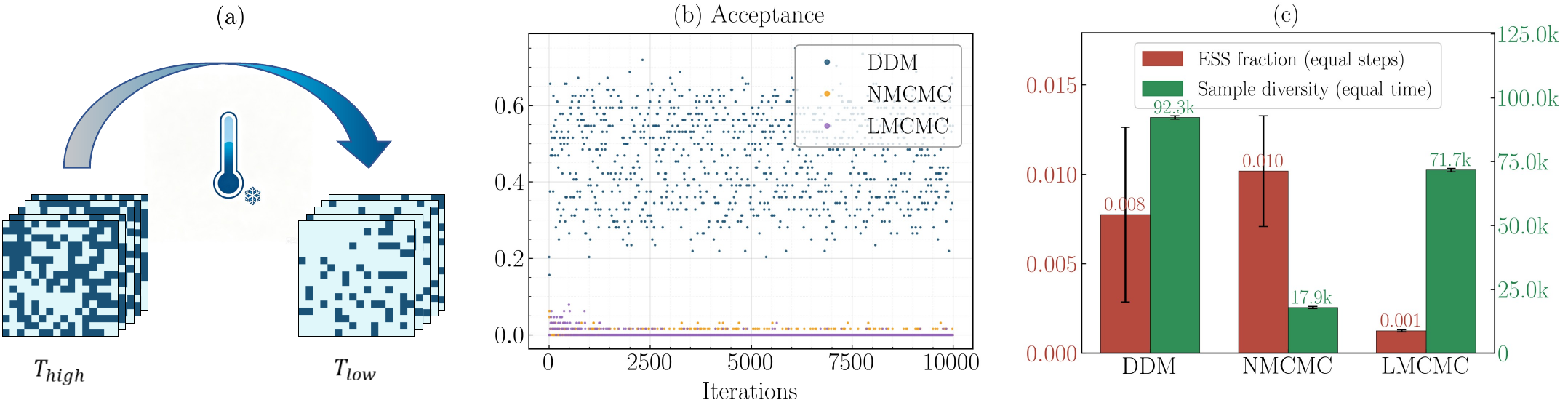}
    \caption{\textbf{Monte Carlo quench.} (a) Schematic of the quench process, where spin lattice configurations transition from higher-temperature initial states to lower-temperature target states. (b) Acceptance rates over 10 000 sampling iterations (data points displayed at 10-iteration intervals), evaluated across 64 independent configurations of the $\mathcal{D} = 16\times16$ 2D Ising model undergoing a quench from $T = 2$ to $T = 1.5$. Three sampling schemes are compared: DDM (discrete diffusion model; connected updates; adaptive acceptance target 0.5), NMCMC (neural Markov chain Monte Carlo; disconnected updates), and LMMC (local Metropolis Monte Carlo). (c) Comparison of magnetization‑based effective sample size (ESS) fraction and sample diversity, where the ESS fraction estimates the proportion of statistically independent samples in MCMC chains and sample diversity quantifies the number of distinct spin configurations explored during sampling. Simulations adopt the same parameters as panel (b), and results are averaged over $10$ independent runs.}
    \label{mc quench}
\end{figure*}

In Fig.~\ref{mc quench}, we present the performance of all sampling methods considered under Monte Carlo “quench” protocols. Notably, equipped with non-local update mechanisms realized via connected and disconnected update schemes, our two proposed methods sustain acceptance rates higher than LMMC throughout the low‑temperature regime, especially the connected update scheme. A naive strategy to break the locality barrier of LMMC is to flip multiple random spins per iteration; unfortunately, this simple modification progressively reduces the acceptance rate. To further quantify the quality of the samples about different methods, we evaluated two quantities in Fig.~\ref{mc quench}(c): the magnetization‑based ESS fraction and sample diversity. The ESS fraction is \(1\big/\left(1+2\sum_{k=1}^{K}\rho(k)\right)\), where \(\rho(k)\) denotes the autocorrelation of the averaged absolute‑magnetization series at lag $k$ and $K$ is the lag at which \(\rho(k)\) first becomes non‑positive. Sample diversity measures the number of distinct spin configurations among the samples. All three methods are run for identical durations, and we compare the diversity of an equal number of samples collected within this fixed runtime. Since NMCMC and LMMC generate more total samples in the same runtime, we randomly take a subset of their samples so that the sample count used for comparison equals that of DDM. For a fair comparison, each method is run on the faster of our available CPU and GPU: DDM and NMCMC on a single NVIDIA GeForce RTX 4090 GPU (24\,GB, CUDA 12.2, PyTorch 2.8), and LMMC on the CPU (AMD Ryzen 9 7950X3D, 16 cores).

In general, our methods explore a substantially broader configuration space than the single-spin-flip update of LMMC while attaining higher acceptance rates. This generates more diverse samples and reduces the total number of iterations required for convergence, at the cost of increased computational time per iteration. Even so, non-local sampling strategies of this kind exhibit promising prospects for efficient equilibrium sampling.

Lastly, we outline three key limitations and practical constraints of our framework. First, the efficiency and reliability of the sampling depend heavily on how well the neural networks are trained, which may fail to capture low‑probability configurations that nevertheless bear physical significance. Second, our framework brings higher computational costs than standard spin-flip update algorithms, since generating candidate samples via the discrete diffusion model and computing their likelihood increases the cost of every iteration, although faster chain mixing can partially compensate for this extra overhead. Third, it remains challenging to establish strict theoretical bounds for the entire MCMC process. 

\section{Discussion}
In this work, we have constructed a new type of discrete diffusion model in which the normalized distributions at successive diffusion times are parameterized by an evolving VAN. Compared with prior discrete diffusion models based on tensor networks \cite{causer2025discrete}, our framework shows favorable performance on high-dimensional lattice systems and alleviates the dimensionality bottleneck commonly encountered by MPS. We incorporate the tau-leaping algorithm to accelerate the sequential denoising process. Relative to the step-by-step denoising baseline, the tau‑leaping‑accelerated method reduces iterative computational overhead while maintaining acceptable numerical precision for most sampling scenarios, achieving a better trade-off between computational cost and sampling accuracy. Comprehensive numerical experiments on both two- and three-dimensional Ising models verify that our method can accurately capture thermodynamic observables including free energy per site, energy per site, and magnetization per site across diverse temperatures, delivering competitive or even better numerical accuracy than tensor-network baselines, especially for challenging three-dimensional spin systems.

We have also shown how to integrate the discrete diffusion model with MCMC to establish an efficient sampling paradigm for discrete spin systems. Our unified pipeline supports two complementary sampling strategies, namely connected update and disconnected update, which together alleviate the local trapping problem that plagues conventional LMMC methods. Specifically, the connected update constructs correlated candidate configurations through sequential forward noising and reverse denoising trajectories. By contrast, the disconnected update performed via NMCMC generates fully independent proposals through VAN sampling. Interestingly, the connected update bears close connections to both LMMC and the disconnected update. Within the connected-update framework, the forward noising process can be interpreted as a sequence of single-spin flips in LMMC, while the reverse denoising process acts as a weakened variant of the disconnected update and recovers the full disconnected update in the limit \(t\rightarrow\infty\).

Finally, we discuss potential extensions of our work for future research. First, the current framework can be extended to more complex spin systems to verify the generalization ability of our diffusion model beyond classical ferromagnetic Ising models. Second, in addition to the tau-leaping adopted in this paper, advanced acceleration techniques such as diffusion model distillation and enhanced tau-leaping schemes \cite{ren2025fast, yao2026acceleratingdiscretediffusionmodels} can be incorporated to accelerate the denoising process. Third, efficient numerical simulation of such stochastic dynamical processes can be extended to advance quantum generative models~\cite{s6zj-vzdp}. Overall, our framework explicitly captures the full distribution dynamics of diffusion, which suggests preliminary directions for subsequent work on model interpretability and architectural optimization.

\section*{Acknowledgments}
We acknowledge Online Club Nanothermodynamica for helpful discussions. This work is supported by Project 12322501, 12575035 of National Natural Science Foundation of China, and 2026NSFSCZY0124 of the Natural Science Foundation of Sichuan Province. 
The HPC is supported by the Center for HPC at University of Electronic Science and Technology of China.

\section*{Data availability}
The authors declare that the data supporting this study are available within the paper.
A PyTorch code implementation of the present algorithm is openly available on GitHub [\href{https://github.com/cowenp/Discrete-Diffusion-Models-via-Evolving-Variational-Autoregressive-Networks.git}{GitHub Repository}]. 

\appendix

\section{Notation}
The key mathematical notations employed throughout this work are defined in Table~\ref{tab:symbol}.
\begin{table}[H]
\centering
\caption{Explanations of mathematical symbols}
\label{tab:symbol}
\renewcommand{\arraystretch}{1.2} 
\begin{tabular}{l p{7cm}} 
\toprule
\textbf{Symbol} & \textbf{Explanation} \\ 
\midrule
$\mathcal{S}$ & Discrete state space\\
$t$ & Time step\\
$\mathcal{D}$ & System size \\ 
$N$ & Batch size \\
$E$ & Energy function \\
$m$ & Absolute magnetization per site\\
$Z$ & Partition function\\ 
$F$ & Free energy \\
$W$ & Markov generator\\
$w$ & Transition rate\\
$R$ & Escape rate\\
$Q$ & Transition matrix\\
$T$ & Temperature of the system \\
$\beta$ & Inverse temperature $\beta = \frac{1}{T}$ of the system\\  
$\pi(y \mid x)$ & Transition probability from $x$ to $y$\\
$u$ & Noisy (latent) spin configuration\\
$\Delta t$ & Diffusion time step, $\Delta t = t_k - t_{k-1}$\\
$C$ & Correlation coefficient\\
$\sigma_i$ & Spin variable, the direction of the $i$-th spin \\ 
$\boldsymbol{\sigma}$ & Spin configuration \\
\bottomrule
\end{tabular}
\end{table}

\section{Masked autoencoder for distribution estimation}
\begin{figure*}[t]
    \centering
    \includegraphics[width=0.8
    \linewidth]{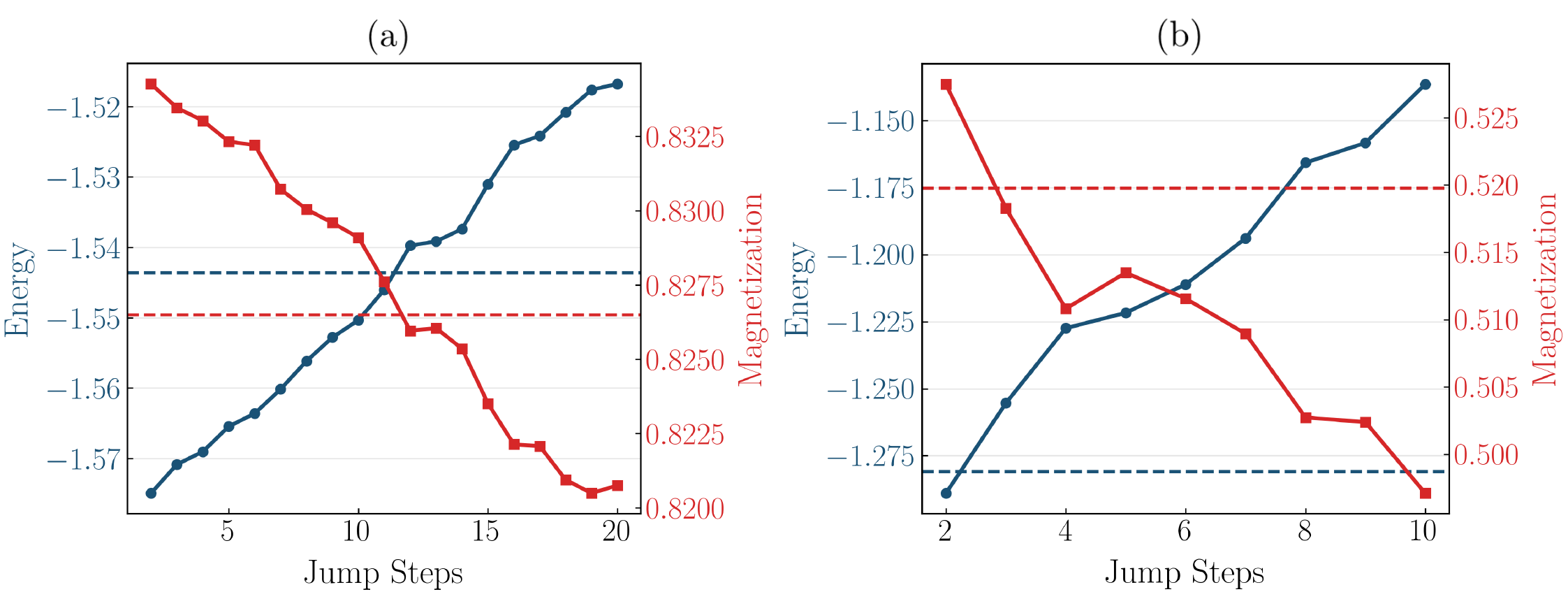}
    \caption{\textbf{Feasibility validation of the tau-leaping method for accelerating the denoising process.} (a) Results for a $\mathcal{D} = 16\times16$ 2D Ising model: we investigate the effects of different tau leaping steps with a fixed total simulation length of 50 denoising steps. Left y-axis shows mean energy per site after denoising as a function of jump step, right y-axis shows mean magnetization per site; solid curves are noised-denoised samples using the tau‑leaping procedure (30 repeats with 200 samples), and dashed horizontal lines indicate baseline values computed from initial samples. (b) Analogous experiment on a 4×4×4 3D Ising model. Together the panels illustrate the behavior of tau-leaping in our discrete diffusion model, and we select the jump step where the solid and dashed curves show the closest agreement as the optimal leap duration.}
    \label{tau-leaping test}
\end{figure*}

In this work, we adopt the masked autoencoder for distribution estimation (MADE) \cite{germain2015made} to model two- and three-dimensional Ising models. For a standard L-layer multi-layer perceptron (MLP), the forward propagation formulation is defined as:
\begin{equation}
\boldsymbol{h}^{l+1} = \sigma\left( \boldsymbol{W}^l \boldsymbol{h}^l + \boldsymbol{b}^l \right)
\end{equation}
where \(\boldsymbol{h}^0=\boldsymbol{x}\) refers to the flattened lattice spin vector, \(\boldsymbol{h}^L\) denotes the final output of the network, and \(\sigma(\cdot)\) is the nonlinear activation function.

To impose autoregressive constraints on the spin lattice configurations, fixed block-wise
binary mask matrices are applied to the weight matrices of all linear layers via element-wise
Hadamard products, which blocks the redundant connection mappings between individual spins.
The forward propagation of the masked layers is then formulated as
\begin{equation}
\boldsymbol{h}^{l+1} = \left( \boldsymbol{W}^l \odot \boldsymbol{M}^l \right)
\sigma\!\left( \boldsymbol{h}^l \right) + \boldsymbol{b}^l ,
\end{equation}
where $\odot$ denotes the element-wise Hadamard product and $\boldsymbol{M}^l$ is a fixed binary mask of the
same shape as $\boldsymbol{W}^l$. Two categories of triangular masks are constructed for different network
layers: the input projection layer adopts an exclusive strictly lower-triangular mask,
$\boldsymbol{T}_{ij}=1$ for $i>j$, which excludes the diagonal and eliminates the
self-dependency of each spin; the remaining masked layers utilize an inclusive
lower-triangular mask, $\boldsymbol{T}_{ij}=1$ for $i\ge j$, which keeps the diagonal and
preserves a valid sequential autoregressive dependency order. Since both patterns are
lower-triangular in the site index, no spin can receive information from a spin of larger
index, which is consistent with the autoregressive factorization.

Regarding network hyper‑parameters, we configure the depth and width conditioned on the lattice dimensions. We set \(\text{net\_depth}=3\) and \(\text{net\_width}=32\) for the two‑dimensional Ising model, while \(\text{net\_depth}=4\) and \(\text{net\_width}=16\) are adopted for the three‑dimensional Ising model. Our VAN is compatible with Adam optimizer \cite{kingma2014adam} and natural gradient descent \cite{liu2025efficient}, which can be incorporated into the parameter update phase to effectively accelerate convergence during model training. In practical training configurations, Adam generally requires a small learning rate, typically set to $10^{-4}$ or $10^{-5}$, while natural gradient descent allows a relatively higher learning rate of $10^{-2}$.

\section{Optimal leap duration for tau-leaping}
\label{Optimal leap}
This appendix explains the principle for determining the optimal leap duration used in the tau-leaping Alg.~\ref{Alg:tau-leaping}, which controls the trade-off between numerical precision and computational efficiency in stochastic simulations of the denoising process. In our discrete diffusion model, we discretize the continuous‑time Markovian dynamics of the forward noising process with fixed‑size jump intervals and adopt a VAN to learn the probability distribution at each time step, which means that the exact transition rates of the denoising process are kept constant within each individual jump interval. In Fig.~\ref{tau-leaping test}, we illustrate the procedure to identify the optimal jump-step size of tau-leaping applied to the denoising process of two- and three-dimensional Ising models. Since the noising process relies on Euler approximation fitting, a greater deviation between the adopted value $\tau$ and the optimal leap duration will lead to lower denoising accuracy. For scenarios that require high precision, it can be difficult to select the appropriate $\tau$ to keep the final results within an acceptable accuracy range. When we do not find a suitable \(\tau\) to keep numerical errors within acceptable limits, Euler-discretized denoising serves as a reliable alternative.

\section{Sampling algorithm}
This appendix presents the full pseudocodes of multiple baseline and reference sampling algorithms discussed in the main text, including the standard local Metropolis Monte Carlo (LMMC), the neural Markov chain Monte Carlo (NMCMC), and the Wolff cluster algorithm. The Wolff cluster algorithm, proposed in 1989 \cite{wolff1989collective}, is a highly efficient single-cluster Monte Carlo method. Its single-cluster update distinguishes it from the closely related Swendsen--Wang (SW) algorithm \cite{swendsen1987nonuniversal}. The primary advantage of the Wolff algorithm is its ability to suppress critical slowing down near second-order phase transitions. In traditional algorithms such as Metropolis sampling, autocorrelation times increase rapidly near phase transitions, drastically reducing sampling efficiency. Samples generated by the Wolff cluster algorithm are used to compute mean energy and magnetization per site, which serve as baseline values for comparison in this work.

Despite its efficiency for critical sampling, the Wolff cluster algorithm exhibits inherent limitations that restrict its general applicability across diverse spin systems and temperature regimes. First, its underlying cluster-building principle is theoretically tailored to ferromagnetic systems with uniform short-range couplings. Second, the cluster construction strategy intrinsically relies on specific symmetry properties and structured spin interactions, which further limits its adaptability to generic disordered or asymmetric spin systems.

\begin{algorithm}[H]
\caption{\centering Wolff Cluster Algorithm}
\label{Wolff Cluster}
\textbf{Step 1.} Initialize the cluster and stack: 

\begin{itemize}
    \item Uniformly select a random lattice site $i$.
    \item Initialize the cluster set: $\mathcal{C} = \{i\}$.
    \item Create an empty stack and push site $i$ onto the stack.
\end{itemize}

\textbf{Step 2.} Grow the cluster using the stack: 

\begin{itemize}
    \item While the stack is not empty, pop a site $j$ from the stack.
    \item Traverse each nearest neighbor $k \in U_j$ of site $j$.
    \item For each neighbor $k$ satisfying $\boldsymbol{\sigma}_k = \boldsymbol{\sigma}_j$ and $k \notin \mathcal{C}$:
          \begin{itemize}
              \item Add $k$ to the cluster $\mathcal{C}$ with probability 
              \[
              p_k = 1 - e^{-2\beta},
              \]
              \item Push the successfully added site $k$ into the stack.
          \end{itemize}
\end{itemize}

\textbf{Step 3.} Flip all spins in the cluster: 
\[
\boldsymbol{\sigma}_j \rightarrow -\boldsymbol{\sigma}_j \quad \forall j \in \mathcal{C}
\]
\end{algorithm}

\begin{algorithm}[H]
\caption{\centering Standard local Metropolis Monte Carlo (LMMC)}
\label{LMMC}
\textbf{Step 1.} Propose a new configuration by flipping the spin of a randomly chosen site:
    \[
    \boldsymbol{\sigma}_i \rightarrow -\boldsymbol{\sigma}_i;
    \]

\textbf{Step 2.} Calculate the energy difference between the new and current configurations:
    \[
    \Delta E = E(\boldsymbol{\sigma}') - E(\boldsymbol{\sigma})
    \]

\textbf{Step 3.} Accept the move with probability:
    \[
    \text{Acc}[\boldsymbol{\sigma} \rightarrow \boldsymbol{\sigma}'] = \min \left[ 1, e^{-\beta \Delta E} \right],
    \]
    Accept $\boldsymbol{\sigma}'$ as the new configuration $\boldsymbol{\sigma}(t+1)=\boldsymbol{\sigma}'$ if the transition is accepted; otherwise, keep the original state $\boldsymbol{\sigma}(t+1) = \boldsymbol{\sigma}$ for the next iteration.
\end{algorithm}

\begin{algorithm}[H]
\caption{\centering Neural Markov Chain Monte Carlo (NMCMC)}
\label{NMCMC}
\textbf{Step 1.} Initialize the lattice spin configuration $\boldsymbol{\sigma}$ and adopt a fixed pre-trained neural generator $P_0^\theta(\cdot)$.

\textbf{Step 2.} Propose a fully independent candidate spin configuration $\boldsymbol{\sigma}'$ directly sampled from the neural generative distribution:
    \[
    \boldsymbol{\sigma}' \sim P_0^\theta(\boldsymbol{\sigma}').
    \]

\textbf{Step 3.} Evaluate the Metropolis–Hastings acceptance probability based on the neural prior distribution and Boltzmann weight to satisfy detailed balance:
    \[
    \text{Acc}[\boldsymbol{\sigma}\to\boldsymbol{\sigma}'] = \min\left[1,\, \frac{P_0^\theta(\boldsymbol{\sigma}) \cdot e^{-\beta E(\boldsymbol{\sigma}')}}{P_0^\theta(\boldsymbol{\sigma}') \cdot e^{-\beta E(\boldsymbol{\sigma})}}\right].
    \]
    Accept $\boldsymbol{\sigma}'$ as the new configuration $\boldsymbol{\sigma}(t+1)=\boldsymbol{\sigma}'$ if the transition is accepted; otherwise, keep the original state $\boldsymbol{\sigma}(t+1) = \boldsymbol{\sigma}$ for the next iteration.
\end{algorithm}

\bibliography{bib}

@inproceedings{sohl2015deep,
  title={Deep unsupervised learning using nonequilibrium thermodynamics},
  author={Sohl-Dickstein, Jascha and Weiss, Eric and Maheswaranathan, Niru and Ganguli, Surya},
  booktitle={International conference on machine learning},
  pages={2256--2265},
  year={2015},
  organization={pmlr},
  url = {https://proceedings.mlr.press/v37/sohl-dickstein15.html}
}

@article{gruver2023protein,
  title={Protein design with guided discrete diffusion},
  author={Gruver, Nate and Stanton, Samuel and Frey, Nathan and Rudner, Tim GJ and Hotzel, Isidro and Lafrance-Vanasse, Julien and Rajpal, Arvind and Cho, Kyunghyun and Wilson, Andrew G},
  journal={Adv. Neural Inf. Process. Syst.},
  volume={36},
  pages={12489--12517},
  year={2023},
  url={https://proceedings.neurips.cc/paper_files/paper/2023/hash/29591f355702c3f4436991335784b503-Abstract-Conference.html}
}

@inproceedings{sanokowski2025scalable,
  title={Scalable discrete diffusion samplers: Combinatorial optimization and statistical physics},
  author={Sanokowski, Sebastian and Berghammer, Wilhelm and Wang, Haoyu and Ennemoser, Martin and Hochreiter, Sepp and Lehner, Sebastian},
  booktitle={International Conference on Learning Representations},
  volume={2025},
  pages={87053--87082},
  year={2025},
  url = {https://proceedings.iclr.cc/paper_files/paper/2025/hash/d87eadbe6114b768e76c5d5e8eb39388-Abstract-Conference.html}
}

@article{singh2026independent,
  title={From Independent to Correlated Diffusion: Generalized Generative Modeling with Probabilistic Computers},
  author={Singh, Nihal Sanjay and Mohseni-Rajaee, Mazdak and Niazi, Shaila and Camsari, Kerem Y},
  journal={arXiv:2603.27996},
  year={2026},
  url={https://arxiv.org/abs/2603.27996}
}

@article{causer2025discrete,
  title={Discrete generative diffusion models without stochastic differential equations: A tensor network approach},
  author={Causer, Luke and Rotskoff, Grant M and Garrahan, Juan P},
  journal={Phys. Rev. E},
  volume={111},
  number={2},
  pages={025302},
  year={2025},
  publisher={APS},
  url={https://doi.org/10.1103/PhysRevE.111.025302}
}

@article{ho2020denoising,
  title={Denoising diffusion probabilistic models},
  author={Ho, Jonathan and Jain, Ajay and Abbeel, Pieter},
  journal={Adv. Neural Inf. Process. Syst.},
  volume={33},
  pages={6840--6851},
  year={2020},
  url = {https://proceedings.neurips.cc/paper/2020/hash/4c5bcfec8584af0d967f1ab10179ca4b-Abstract.html}
}

@article{austin2021structured,
  title={Structured denoising diffusion models in discrete state-spaces},
  author={Austin, Jacob and Johnson, Daniel D and Ho, Jonathan and Tarlow, Daniel and Van Den Berg, Rianne},
  journal={Adv. Neural Inf. Process. Syst.},
  volume={34},
  pages={17981--17993},
  year={2021},
  url={https://proceedings.neurips.cc/paper/2021/hash/6f558439a07a94918a07773f523f87a-Abstract.html}
}

@article{lou2023discrete,
  title={Discrete diffusion modeling by estimating the ratios of the data distribution},
  author={Lou, Aaron and Meng, Chenlin and Ermon, Stefano},
  journal={arXiv:2310.16834},
  year={2023},
  url={https://arxiv.org/abs/2310.16834}
}

@inproceedings{arriola2025block,
  title={Block diffusion: Interpolating between autoregressive and diffusion language models},
  author={Arriola, Marianne and Gokaslan, Aaron and Chiu, Justin and Yang, Zhihan and Qi, Zhixuan and Han, Jiaqi and Sahoo, Subham and Kuleshov, Volodymyr},
  booktitle={International Conference on Learning Representations},
  volume={2025},
  pages={50726--50753},
  year={2025},
  url = {https://proceedings.iclr.cc/paper_files/paper/2025/hash/7ede97c3e082c6df10a8d6103a2eebd2-Abstract-Conference.html}
}

@article{batzolis2021conditional,
  title={Conditional image generation with score-based diffusion models},
  author={Batzolis, Georgios and Stanczuk, Jan and Sch{\"o}nlieb, Carola-Bibiane and Etmann, Christian},
  journal={arXiv:2111.13606},
  year={2021},
  url = {https://arxiv.org/abs/2111.13606}
}

@article{ho2022cascaded,
  title={Cascaded diffusion models for high fidelity image generation},
  author={Ho, Jonathan and Saharia, Chitwan and Chan, William and Fleet, David J and Norouzi, Mohammad and Salimans, Tim},
  journal={J. Mach. Learn. Res.},
  volume={23},
  number={47},
  pages={1--33},
  year={2022},
  url = {https://jmlr.org/papers/volume23/21-0635/21-0635.pdf}
}

@article{schneider2023archisound,
  title={Archisound: Audio generation with diffusion},
  author={Schneider, Flavio},
  journal={arXiv:2301.13267},
  year={2023},
  url = {https://arxiv.org/abs/2301.13267}
}

@article{del2025performance,
  title={Performance of machine-learning-assisted Monte Carlo in sampling from simple statistical physics models},
  author={Del Bono, Luca Maria and Ricci-Tersenghi, Federico and Zamponi, Francesco},
  journal={Phys. Rev. E},
  volume={112},
  number={4},
  pages={045307},
  year={2025},
  publisher={APS},
  url={https://doi.org/10.1103/PhysRevE.112.045307}
}

@article{hunt2024accelerating,
  title={Accelerating Markov chain Monte Carlo sampling with diffusion models},
  author={Hunt-Smith, Nicholas T and Melnitchouk, Wally and Ringer, Felix and Sato, Nobuo and Thomas, Anthony W and White, Martin J},
  journal={Comput. Phys. Commun.},
  volume={296},
  pages={109059},
  year={2024},
  publisher={Elsevier},
  url = {https://www.sciencedirect.com/science/article/pii/S0010465523004046?via%3Dihub}
}

@article{ren2025fast,
  title={Fast solvers for discrete diffusion models: Theory and applications of high-order algorithms},
  author={Ren, Yinuo and Chen, Haoxuan and Zhu, Yuchen and Guo, Wei and Chen, Yongxin and Rotskoff, Grant M and Tao, Molei and Ying, Lexing},
  journal={arXiv:2502.00234},
  year={2025},
  url={https://arxiv.org/abs/2502.00234}
}

@article{campbell2022continuous,
  title={A continuous time framework for discrete denoising models},
  author={Campbell, Andrew and Benton, Joe and De Bortoli, Valentin and Rainforth, Thomas and Deligiannidis, George and Doucet, Arnaud},
  journal={Adv. Neural Inf. Process. Syst.},
  volume={35},
  pages={28266--28279},
  year={2022},
  url={https://papers.nips.cc/paper_files/paper/2022/hash/b5b528767aa35f5b1a60fe0aaeca0563-Abstract-Conference.html}
}

@article{wu2019solving,
  title={Solving statistical mechanics using variational autoregressive networks},
  author={Wu, Dian and Wang, Lei and Zhang, Pan},
  journal={Phys. Rev. Lett.},
  volume={122},
  number={8},
  pages={080602},
  year={2019},
  publisher={APS},
  url={https://doi.org/10.1103/PhysRevLett.122.080602}
}

@article{gillespie2001approximate,
  title={Approximate accelerated stochastic simulation of chemically reacting systems},
  author={Gillespie, Daniel T},
  journal={J. Chem. Phys.},
  volume={115},
  number={4},
  pages={1716--1733},
  year={2001},
  publisher={AIP Publishing},
  url={https://pubs.aip.org/aip/jcp/article-abstract/115/4/1716/451187/Approximate-accelerated-stochastic-simulation-of?redirectedFrom=fulltext}
}

@article{cao2006efficient,
  title={Efficient step size selection for the tau-leaping simulation method},
  author={Cao, Yang and Gillespie, Daniel T and Petzold, Linda R},
  journal={J. Chem. Phys.},
  volume={124},
  number={4},
  year={2006},
  publisher={AIP Publishing},
  url={https://pubs.aip.org/aip/jcp/article-abstract/124/4/044109/562210/Efficient-step-size-selection-for-the-tau-leaping?redirectedFrom=fulltext}
}

@article{cao2005avoiding,
  title={Avoiding negative populations in explicit Poisson tau-leaping},
  author={Cao, Yang and Gillespie, Daniel T and Petzold, Linda R},
  journal={J. Chem. Phys.},
  volume={123},
  number={5},
  year={2005},
  publisher={AIP Publishing},
  url={https://pubs.aip.org/aip/jcp/article-abstract/123/5/054104/905859/Avoiding-negative-populations-in-explicit-Poisson?redirectedFrom=fulltext}
}

@article{gillespie1976general,
  title={A general method for numerically simulating the stochastic time evolution of coupled chemical reactions},
  author={Gillespie, Daniel T},
  journal={J. Comput. Phys.},
  volume={22},
  number={4},
  pages={403--434},
  year={1976},
  publisher={Elsevier},
  url={http://web.mit.edu/endy/www/scraps/signal/JCompPhys%2822%29403.pdf}
}

@article{gillespie1977exact,
  title={Exact stochastic simulation of coupled chemical reactions},
  author={Gillespie, Daniel T},
  journal={J. Phys. Chem.},
  volume={81},
  number={25},
  pages={2340--2361},
  year={1977},
  publisher={ACS Publications},
  url={http://web.mit.edu/endy/www/scraps/dg/JPC%2881%292340.pdf}
}

@article{tang2024learning,
  title={Learning nonequilibrium statistical mechanics and dynamical phase transitions},
  author={Tang, Ying and Liu, Jing and Zhang, Jiang and Zhang, Pan},
  journal={Nat. Commun.},
  volume={15},
  number={1},
  pages={1117},
  year={2024},
  publisher={Nature Publishing Group UK London},
  url={https://www.nature.com/articles/s41467-024-45172-8}
}

@article{liu2025efficient,
  title={Efficient optimization of variational autoregressive networks with natural gradient},
  author={Liu, Jing and Tang, Ying and Zhang, Pan},
  journal={Phys. Rev. E},
  volume={111},
  number={2},
  pages={025304},
  year={2025},
  publisher={APS},
  url = {https://arxiv.org/abs/2409.20029}
}

@article{yu2025discrete,
  title={Discrete diffusion in large language and multimodal models: A survey},
  author={Yu, Runpeng and Li, Qi and Wang, Xinchao},
  journal={arXiv:2506.13759},
  year={2025},
  url = {https://arxiv.org/abs/2506.13759}
}

@inproceedings{desurvey,
  title={A Survey of Discrete Diffusion for Text and Genomic Sequence Generation},
  author={de Groot, Lars and Kuiper, Ruurd Jan Anthonius and Bagheri, Ayoub},
  booktitle={The 37th Benelux Conference on Artificial Intelligence and the 34th Belgian Dutch Conference on Machine Learning},
  url = {https://openreview.net/forum?id=ubOzVt7oMw}
}

@article{zhu2026mdns,
  title={Mdns: Masked diffusion neural sampler via stochastic optimal control},
  author={Zhu, Yuchen and Guo, Wei and Choi, Jaemoo and Liu, Guan-Horng and Chen, Yongxin and Tao, Molei},
  journal={Adv. Neural Inf. Process. Syst.},
  volume={38},
  pages={35260--35308},
  year={2025},
  url = {https://papers.neurips.cc/paper_files/paper/2025/hash/3289cc9cb9fdde172004497c61359544-Abstract-Conference.html}
}

@article{kathuria2026leveraging,
  title={Leveraging Pretrained Language Models as Energy Functions for Glauber Dynamics Text Diffusion},
  author={Kathuria, Tarun and Kumar, Sachin},
  journal={arXiv:2605.04291},
  year={2026},
  url = {https://arxiv.org/abs/2605.04291}
}

@book{chandler1987introduction,
  title={Introduction to Modern Statistical Mechanics},
  author={Chandler, David},
  year={1987},
  publisher={Oxford University Press},
  address={Oxford, UK}
}

@article{wolff1989collective,
  title={Collective Monte Carlo updating for spin systems},
  author={Wolff, Ulli},
  journal={Phys. Rev. Lett.},
  volume={62},
  number={4},
  pages={361},
  year={1989},
  publisher={APS},
  url = {https://link.aps.org/doi/10.1103/PhysRevLett.62.361}
}

@article{swendsen1987nonuniversal,
  title={Nonuniversal critical dynamics in Monte Carlo simulations},
  author={Swendsen, Robert H and Wang, Jian-Sheng},
  journal={Phys. Rev. Lett.},
  volume={58},
  number={2},
  pages={86},
  year={1987},
  publisher={APS},
  url = {https://journals.aps.org/prl/abstract/10.1103/PhysRevLett.58.86}
}

@inproceedings{wang2025fine,
  title={Fine-tuning discrete diffusion models via reward optimization with applications to dna and protein design},
  author={Wang, Chenyu and Uehara, Masatoshi and He, Yichun and Wang, Amy and Lal, Avantika and Jaakkola, Tommi and Levine, Sergey and Regev, Aviv and Wang, Hanchen and Biancalani, Tommaso},
  booktitle={International Conference on Learning Representations},
  volume={2025},
  pages={47871--47899},
  year={2025},
  url = {https://proceedings.iclr.cc/paper_files/paper/2025/hash/771e09dd204ea339da0d8114c48afd21-Abstract-Conference.html}
}

@article{sarkar2024designing,
  title={Designing DNA with tunable regulatory activity using discrete diffusion},
  author={Sarkar, Anirban and Tang, Ziqi and Zhao, Chris and Koo, Peter K},
  journal={bioRxiv},
  pages={2024--05},
  year={2024},
  publisher={Cold Spring Harbor Laboratory},
  url = {https://www.biorxiv.org/content/10.1101/2024.05.23.595630v1}
}

@article{levin2007tensor,
  title={Tensor renormalization group approach to two-dimensional classical lattice models},
  author={Levin, Michael and Nave, Cody P},
  journal={Phys. Rev. Lett.},
  volume={99},
  number={12},
  pages={120601},
  year={2007},
  publisher={APS},
  url = {https://journals.aps.org/prl/abstract/10.1103/PhysRevLett.99.120601}
}

@article{xie2009second,
  title={Second renormalization of tensor-network states},
  author={Xie, Zhi-Yuan and Jiang, Hong-Chen and Chen, Qinjun N and Weng, Zheng-Yu and Xiang, Tao},
  journal={Phys. Rev. Lett.},
  volume={103},
  number={16},
  pages={160601},
  year={2009},
  publisher={APS},
  url = {https://journals.aps.org/prl/abstract/10.1103/PhysRevLett.103.160601}
}

@inproceedings{germain2015made,
  title={Made: Masked autoencoder for distribution estimation},
  author={Germain, Mathieu and Gregor, Karol and Murray, Iain and Larochelle, Hugo},
  booktitle={International conference on machine learning},
  pages={881--889},
  year={2015},
  organization={PMLR},
  url = {https://proceedings.mlr.press/v37/germain15.html}
}

@article{mcnaughton2020boosting,
  title={Boosting Monte Carlo simulations of spin glasses using autoregressive neural networks},
  author={McNaughton, B and Milo{\v{s}}evi{\'c}, MV and Perali, A and Pilati, S},
  journal={Phys. Rev. E},
  volume={101},
  number={5},
  pages={053312},
  year={2020},
  publisher={APS},
  url = {https://journals.aps.org/pre/abstract/10.1103/PhysRevE.101.053312}
}

@article{biazzo2024sparse,
  title={Sparse autoregressive neural networks for classical spin systems},
  author={Biazzo, Indaco and Wu, Dian and Carleo, Giuseppe},
  journal={Mach. Learn.: Sci. Technol.},
  volume={5},
  number={2},
  pages={025074},
  year={2024},
  publisher={IOP Publishing},
  url = {https://iopscience.iop.org/article/10.1088/2632-2153/ad5783}
}

@article{mehta2019high,
  title={A high-bias, low-variance introduction to machine learning for physicists},
  author={Mehta, Pankaj and Bukov, Marin and Wang, Ching-Hao and Day, Alexandre GR and Richardson, Clint and Fisher, Charles K and Schwab, David J},
  journal={Phys. Rep.},
  volume={810},
  pages={1--124},
  year={2019},
  publisher={Elsevier},
  url = {https://www.sciencedirect.com/science/article/pii/S0370157319300766}
}

@article{wang2026commit,
  title={When to Commit? Towards Variable-Size Self-Contained Blocks for Discrete Diffusion Language Models},
  author={Wang, Danny and Qiu, Ruihong and Huang, Zi},
  journal={arXiv:2604.23994},
  year={2026},
  url = {https://arxiv.org/abs/2604.23994}
}

@article{zhou2026steering,
  title={Steering Without Breaking: Mechanistically Informed Interventions for Discrete Diffusion Language Models},
  author={Zhou, Hanhan and Roy, Shamik and Gangadharaiah, Rashmi},
  journal={arXiv:2605.10971},
  year={2026},
  url = {https://arxiv.org/abs/2605.10971}
}

@incollection{ketkar2021convolutional,
  title={Convolutional neural networks},
  author={Ketkar, Nikhil and Moolayil, Jojo},
  booktitle={Deep learning with Python: learn best practices of deep learning models with PyTorch},
  pages={197--242},
  year={2021},
  publisher={Springer},
  url = {https://link.springer.com/book/10.1007/978-1-4842-5364-9}
}

@article{sun2022score,
  title={Score-based continuous-time discrete diffusion models},
  author={Sun, Haoran and Yu, Lijun and Dai, Bo and Schuurmans, Dale and Dai, Hanjun},
  journal={arXiv:2211.16750},
  year={2022},
  url = {https://arxiv.org/abs/2211.16750}
}

@article{del2026demonstrating,
  title={Demonstrating real advantage of machine learning--enhanced Monte Carlo for combinatorial optimization},
  author={Del Bono, Luca Maria and Ricci-Tersenghi, Federico and Zamponi, Francesco},
  journal={Proc. Natl. Acad. Sci.},
  volume={123},
  number={19},
  pages={e2534768123},
  year={2026},
  publisher={National Academy of Sciences},
  url = {https://www.pnas.org/doi/10.1073/pnas.2534768123}
}

@article{de2021diffusion,
  title={Diffusion schr{\"o}dinger bridge with applications to score-based generative modeling},
  author={De Bortoli, Valentin and Thornton, James and Heng, Jeremy and Doucet, Arnaud},
  journal={Adv. Neural Inf. Process. Syst.},
  volume={34},
  pages={17695--17709},
  year={2021},
  url = {https://proceedings.neurips.cc/paper_files/paper/2021/file/940392f5f32a7ade1cc201767cf83e31-Paper.pdf}
}

@article{yang2023diffusion,
  title={Diffusion models: A comprehensive survey of methods and applications},
  author={Yang, Ling and Zhang, Zhilong and Song, Yang and Hong, Shenda and Xu, Runsheng and Zhao, Yue and Zhang, Wentao and Cui, Bin and Yang, Ming-Hsuan},
  journal={ACM Comput. Surv.},
  volume={56},
  number={4},
  pages={1--39},
  year={2023},
  publisher={ACM New York, NY, USA},
  url = {https://dl.acm.org/doi/pdf/10.1145/3626235}
}

@article{bahri2020statistical,
  title={Statistical mechanics of deep learning},
  author={Bahri, Yasaman and Kadmon, Jonathan and Pennington, Jeffrey and Schoenholz, Sam S and Sohl-Dickstein, Jascha and Ganguli, Surya},
  journal={Annu. Rev. Condens. Matt. Phys.},
  volume={11},
  number={1},
  pages={501--528},
  year={2020},
  publisher={Annual Reviews},
  url = {https://www.annualreviews.org/content/journals/10.1146/annurev-conmatphys-031119-050745}
}

@article{lemercier2025diffusion,
  title={Diffusion models for audio restoration: A review},
  author={Lemercier, Jean-Marie and Richter, Julius and Welker, Simon and Moliner, Eloi and V{\"a}lim{\"a}ki, Vesa and Gerkmann, Timo},
  journal={IEEE Signal Processing Magazine},
  volume={41},
  number={6},
  pages={72--84},
  year={2025},
  publisher={IEEE},
  url = {https://ieeexplore.ieee.org/document/10819705}
}

@article{tang2023neural,
  title={Neural-network solutions to stochastic reaction networks},
  author={Tang, Ying and Weng, Jiayu and Zhang, Pan},
  journal={Nat. Mach. Intell.},
  volume={5},
  number={4},
  pages={376--385},
  year={2023},
  publisher={Nature Publishing Group UK London},
  url = {https://www.nature.com/articles/s42256-023-00632-6}
}

@article{weng2025tracking,
  title={Tracking large chemical reaction networks and rare events by neural networks},
  author={Weng, Jiayu and Zhu, Xinyi and Liu, Jing and L{\"u}, Linyuan and Zhang, Pan and Tang, Ying},
  journal={arXiv:2512.10309},
  year={2025},
  url={https://arxiv.org/abs/2512.10309}
}

@article{shin2021protein,
  title={Protein design and variant prediction using autoregressive generative models},
  author={Shin, Jung-Eun and Riesselman, Adam J and Kollasch, Aaron W and McMahon, Conor and Simon, Elana and Sander, Chris and Manglik, Aashish and Kruse, Andrew C and Marks, Debora S},
  journal={Nat. Commun.},
  volume={12},
  number={1},
  pages={2403},
  year={2021},
  publisher={Nature Publishing Group UK London},
  url = {https://www.nature.com/articles/s41467-021-22732-w}
}

@article{carleo2017solving,
  title={Solving the quantum many-body problem with artificial neural networks},
  author={Carleo, Giuseppe and Troyer, Matthias},
  journal={Science},
  volume={355},
  number={6325},
  pages={602--606},
  year={2017},
  publisher={American Association for the Advancement of Science},
  url = {https://www.science.org/doi/10.1126/science.aag2302}
}

@article{sharir2020deep,
  title={Deep autoregressive models for the efficient variational simulation of many-body quantum systems},
  author={Sharir, Or and Levine, Yoav and Wies, Noam and Carleo, Giuseppe and Shashua, Amnon},
  journal={Phys. Rev. Lett.},
  volume={124},
  number={2},
  pages={020503},
  year={2020},
  publisher={APS},
  url = {https://journals.aps.org/prl/abstract/10.1103/PhysRevLett.124.020503}
}

@article{luo2022autoregressive,
  title={Autoregressive neural network for simulating open quantum systems via a probabilistic formulation},
  author={Luo, Di and Chen, Zhuo and Carrasquilla, Juan and Clark, Bryan K},
  journal={Phys. Rev. Lett.},
  volume={128},
  number={9},
  pages={090501},
  year={2022},
  publisher={APS},
  url = {https://journals.aps.org/prl/abstract/10.1103/PhysRevLett.128.090501}
}

@article{kingma2014adam,
  title={Adam: A method for stochastic optimization},
  author={Kingma, Diederik P and Ba, Jimmy},
  journal={arXiv:1412.6980},
  year={2014},
  url={https://arxiv.org/abs/1412.6980},
}

@article{yao2026acceleratingdiscretediffusionmodels,
      title={Accelerating Discrete Diffusion Models with Parallel-In-Time Sampling}, 
      author={Yu Yao and Huanjian Zhou and Andi Han and Wei Huang and Masashi Sugiyama},
      year={2026},
      journal={arXiv:2607.00773},
      url={https://arxiv.org/abs/2607.00773}, 
}

@article{chen2026markov,
  title={Markov Chain Monte Carlo with Diffusion Paths},
  author={Chen, Han and Liu, Sifan and Yang, Jun},
  journal={arXiv:2607.11631},
  year={2026},
  url = {https://arxiv.org/abs/2607.11631}
}

@article{zhong2026scalable,
  title={Scalable Physics-Inspired Transformers for Spin Glasses},
  author={Zhong, Lu and Duan, Wenli and Liu, Jing and Zhang, Pan and Tang, Ying},
  journal={arXiv:2606.22984},
  year={2026},
  url={https://arxiv.org/abs/2606.22984}
}

@article{s6zj-vzdp,
  title = {Quantum Flow Matching},
  author = {Cui, Zidong and Zhang, Pan and Tang, Ying},
  journal = {PRX Intelligence},
  volume = {1},
  issue = {1},
  pages = {013009},
  numpages = {15},
  year = {2026},
  month = {Aug},
  publisher = {American Physical Society},
  doi = {10.1103/s6zj-vzdp},
  url = {https://link.aps.org/doi/10.1103/s6zj-vzdp}
}

\end{document}